\documentclass{article}

\usepackage{PRIMEarxiv}

\usepackage[utf8]{inputenc} 
\usepackage[T1]{fontenc}    
\usepackage{hyperref}       
\usepackage{url}            
\usepackage{booktabs}       
\usepackage{amsfonts}       
\usepackage{nicefrac}       
\usepackage{microtype}      
\usepackage{lipsum}
\usepackage{fancyhdr}       
\usepackage{graphicx}       
\usepackage{wrapfig} 
\graphicspath{{media/}}     
\usepackage{float}

\title{Evaluating Simple Debiasing Techniques in 
RoBERTa-based Hate Speech Detection Models
}

\author{
  Diana Iftimie
  \\ (\texttt{dianai@berkeley.edu}) \And
  Erik Zinn
  \\ (\texttt{erikzinn@berkeley.edu})
}

\begin{document}
\maketitle

\begin{abstract}
The hate speech detection task is known to suffer from bias against African American English (AAE) dialect text, due to the annotation bias present in the underlying hate speech datasets used to train these models. This leads to a disparity where normal AAE text is more likely to be misclassified as abusive/hateful compared to non-AAE text. Simple debiasing techniques have been developed in the past to counter this sort of disparity, and in this work, we apply and evaluate these techniques in the scope of RoBERTa-based encoders. Experimental results suggest that the success of these techniques depends heavily on the methods used for training dataset construction, but with proper consideration of representation bias, they can reduce the disparity seen among dialect subgroups on the hate speech detection task. \textit{Original Report Publication: December 2020}
\end{abstract}

\section{Introduction and Background}
As today’s society relies more and more heavily on the advent of social media, the freedom for people to express their thoughts and beliefs has never been easier. But as discourse shifts to these online platforms, effectively detecting hate speech among the many valid opinions posted to social media has become a prevalent issue. As has been noted time and again, hate speech can truly negate the experience of users and spread false information to the public, and is something that social media companies such as Twitter and Facebook work hard to try to eliminate to ensure their platforms promote positive discourse.

Most of the datasets used to train models for detecting hate speech are collected via crowdsourcing annotations (Davidson 2017 \cite{Davidson}¸ Founta 2018 \cite{Founta}); but as was noted in Sap 2019 \cite{Sap}, many of the models developed from these datasets, while performing well on their target task, are biased against certain minorities, simply because the underlying datasets used to train these models were vulnerable to annotation bias. Specifically, they found that non-expert annotators are more likely to label text as abusive/harmful compared to expert annotators, and that there is a particular disparity between text written in an African-American English (AAE) dialect, compared to a non-AAE dialect. In the process, these models inherit the bias from their datasets, resulting in models that are more likely to identify normal AAE text as abusive, which can result in content moderation strategies that suppress minority groups from being able to have a voice on these online platforms.

Because reannotating existing datasets can be expensive and time-intensive, efforts have been made in recent years to instead “debias” the models against the sensitive attribute of dialect at training time, given the known disparity identified in Sap 2019 \cite{Sap}. Xia 2019 \cite{Xia} showcased an adversarial-based debiasing technique to debias the underlying Bi-LSTM-based hate speech detection model, while Mozafari 2020 \cite{Mozafari} looked at a more complex “bias-fixing mechanism” tuned for BERT-based encoders. Earlier on, Beutel 2017 \cite{Beutel} showcased a general-purpose, gradient negation debiasing technique, applied to the task of predicting income measures, given gender as a sensitive attribute.

While many debiasing techniques have made their way into the NLP literature, very few have tackled the hate speech detection task, and none of them using relatively simple debiasing techniques applied to transformer-based encoders. In this report, we apply the two simple techniques introduced by Xia 2019 \cite{Xia} and Beutel 2017 \cite{Beutel} to the hate speech detection task and evaluate the effectiveness of these debiasing techniques in the context of RoBERTa-based encoders (Liu 2019 \cite{Liu}). We focus our analysis on the dataset from Founta 2018 \cite{Founta}, augmented with dialect labels produced by a demographic-based dialect classifier from Blodgett 2016 \cite{Blodgett}. We explore the impact the training set distribution (with respect to hate speech labels and the dialect subgroups) has on the results of debiasing, and evaluate the performance (with respect to correcting model bias) using traditional metrics and false positive rates, as well as model fairness metrics introduced in Beutel 2017 \cite{Beutel}.  

\section{Data}

\begin{wrapfigure}{r}{0.39\textwidth}
    \centering
    \includegraphics[width=\linewidth]{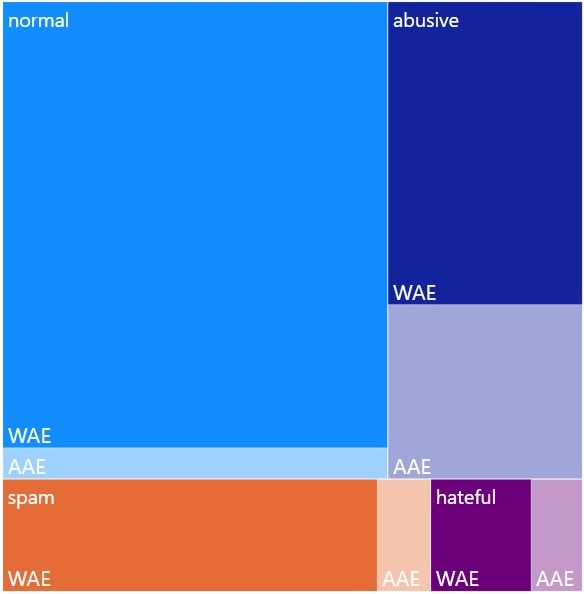}
    \caption{Founta Dataset Tree Map, grouped by Hate Speech classes and Dialect classes\\}
    \label{fig:fig1}

    \centering
    \includegraphics[width=\linewidth]{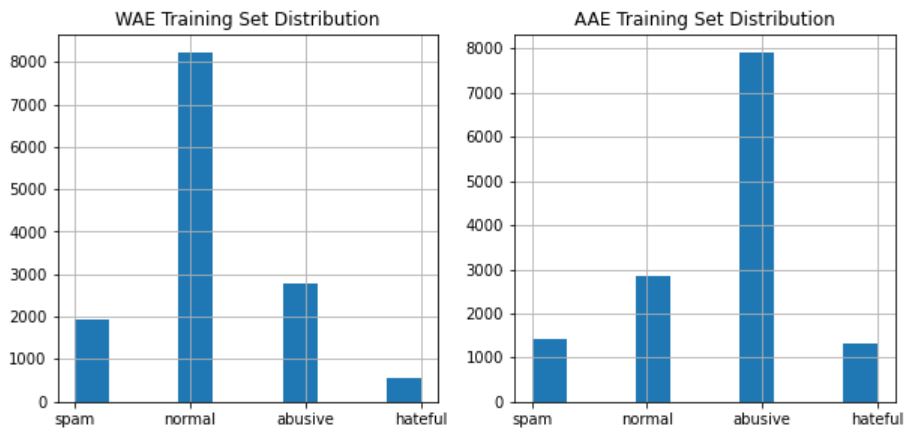}
    \caption{Case 1 - Training Dataset with Representation Bias\\}
    \label{fig:fig2}

    \centering
    \includegraphics[width=\linewidth]{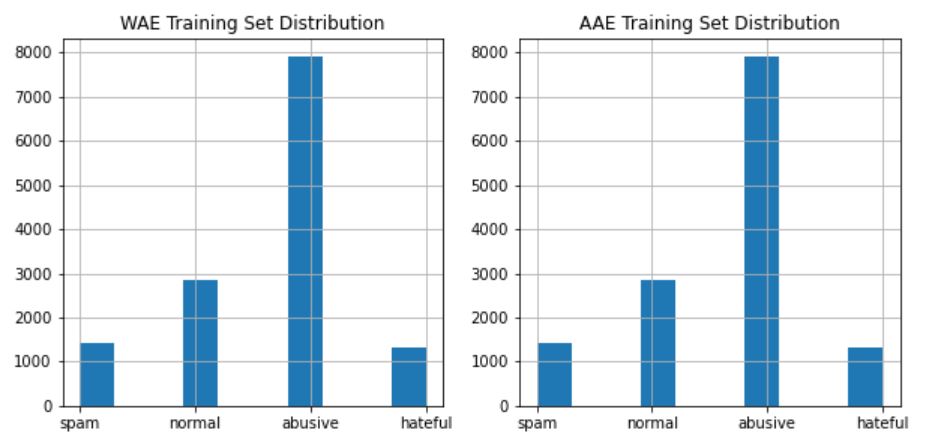}
    \caption{Case 2 - Training Dataset without Representation Bias}
    \label{fig:fig3}
\end{wrapfigure}

For this report, we chose to focus on the dataset collected in Founta 2018 \cite{Founta} (used in both Sap 2019 \cite{Sap} and Xia 2019 \cite{Xia}). This dataset consists of approximately 100,000 records of tweets that were labeled for hate speech using crowd-sourcing techniques. At a high level, this dataset consists of four classes of data—two negative classes of “spam” and “normal”, and two positive classes of “abusive” and “hateful”. 

In consistency with other works, we leverage the demographic-based classifier developed in Blodgett 2016 \cite{Blodgett} to augment the Founta dataset with dialect labels. Given a piece of text, the classifier predicts a posterior probability of four dialect categories of English; but as per the original authors, we chose to only focus on the scores for African-American English (AAE) and White-Aligned English (WAE). Additionally, to be consistent with Sap 2019 \cite{Sap} and Xia 2019 \cite{Xia}, we chose to label tweets with the majority (higher-probability) class between the two dialect classes, resulting in the distribution of hate speech classes and dialect classes for the Founta dataset shown in Figure \ref{fig:fig1}. From the start, there’s an apparent data skew between AAE and WAE data, especially with regard to the negative hate speech classes. This data is split into a train and test subset from the beginning, and we evolve the training subset in later steps.

As discussed in Beutel 2017 \cite{Beutel}, debiasing and model fairness depends heavily on the representation of the training data with respect to the sensitive attribute (in this case, dialect). Thus, we concentrated on choosing a balanced training set with respect to the dialect by undersampling the WAE subset of the Founta data. 

Additionally, Beutel 2017 \cite{Beutel} discusses the implications of representation bias in terms of the target task. 
To examine for representation bias on the target task, we must ask whether the data distribution with respect to the target attribute (hate speech labels) looks similar for both subgroups with respect to the sensitive attribute (dialect). For our purposes, we explored two primary cases: 

\begin{enumerate}
    \item Subgroup representation distributions more closely matching those of the original Founta dataset distributions (Figure \ref{fig:fig2}) 
    \item Subgroup representation distributions that are consistent between dialect subgroups (Figure \ref{fig:fig3})
\end{enumerate}
again by undersampling the WAE subset of the Founta data.

\section{Methodology}

\subsection{Encoder, Tokenizer, and Preprocessing}
Given our goal of applying the aforementioned debiasing techniques to a transformer-based encoder, we chose to use RoBERTa (Liu 2019 \cite{Liu}) as our base encoder, as it is known to have better performance compared to more common encoders such as BERT (Devlin 2019 \cite{Devlin}), although not as fast at training as smaller models such as DistilBERT (Sanh 2019 \cite{Sanh}), which we used for prototyping. Specifically, we used the Tensorflow-based pretrained model provided by HuggingFace, along with the corresponding RobertaTokenizer for preparing our textual embeddings. Prior to running the tokenizer on input text, we drop urls and emojis, replace twitter handles (e.g. "@username") with "user", and keeping hashtags as is.

\subsection{Alternating Adversarial Debiasing Technique}

\begin{wrapfigure}[10]{r}{0.45\linewidth}
\centering
\includegraphics[width=\linewidth]{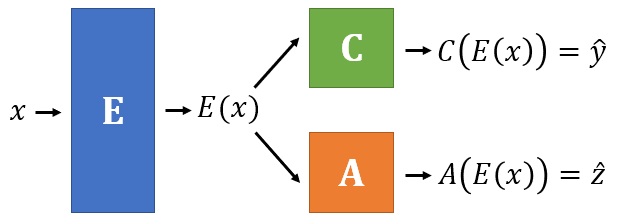}
\caption{High-level model architecture for Alternating Adversarial Debiasing Technique}
\label{fig:fig4}
\end{wrapfigure}

The first debiasing technique we explored adapting was the alternating adversarial technique introduced in Xia 2019 \cite{Xia}. In this technique, the model consists of three components: the encoder $E$, the hate speech classifier $C$ and the dialect classifier (or adversary) $A$. Both the classifier and adversary receive and share the input from the encoder $E$ for doing their individual classification tasks, and are shown in the diagram at right \ref{fig:fig4}, where $x$ is the input text, $y$ is the target variable (hate speech), and $z$ is the sensitive attribute (dialect). Training then consists of following steps: 

\begin{enumerate}
    \item \textbf{Train $C$}: Train the encoder $E$ and classifier $C$ for the hate speech task, minimizing the loss with respect to the target attribute $y$. The adversary $A$ is ignored.
    \item \textbf{Train $A$}: Freeze the encoder $E$ and train the adversary $A$, minimizing the loss with respect to the sensitive attribute $z$, where the objective is:
    $$min_{A} \frac{1}{N}\sum_{i=1}^{N}\mathcal{L}(A(E(x_i)),z_i)$$ This will train the adversary to correctly distinguish the dialect from the encoder output. The classifier $C$ is ignored.
    \item \textbf{Debias $E$}: Freeze the dialect adversary $A$ and unfreeze the encoder $E$ and classifier $C$, minimizing the loss with respect to the target attribute $y$ and the sensitive attribute $z$, where the objective is: 
    $$
    min_{E,C} \frac{1}{N}\sum_{i=1}^{N}\alpha\mathcal{L}(C(E(x_i)),y_i) + (1-\alpha)\mathcal{L}(A(E(x_i)),0.5)
    $$
    This will train the encoder to “fool” the adversary by generating textual representations that will cause the adversary to output random guesses, rather than correct predictions, while training the classifier to correctly predict the target attribute from the encoder output.
    \item Alternate training $A$ and debiasing $E$ by repeating steps 2 and 3 for a total of 10 rounds, where one round consists of applying steps 2 and 3 once.  
\end{enumerate}

\begin{figure}[H]
\centering
\includegraphics[width=\linewidth]{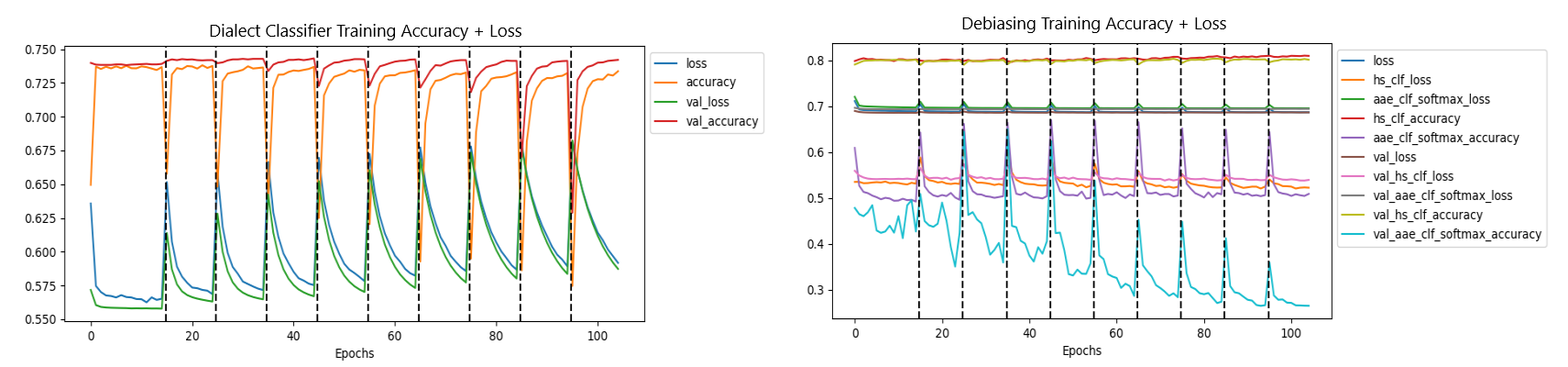}
\caption{Example training progress for Alternating Adversarial Debiasing Technique}
\label{fig:fig5}
\end{figure}

Over time, the adversary should perform poorly at its task after debiasing, with periods of improvement at each execution of step 2 where it attempts to “relearn” how to predict the sensitive attribute. Meanwhile, the classifier should maintain its performance in predicting hate speech in the ideal case. Figure \ref{fig:fig5} showcases the results from training in step 2 (left) and step 3 (right); the results shown in the figure were generated with training data that was balanced with respect to dialect but contains representation bias that was present in the original Founta dataset (Figure \ref{fig:fig2}).

\subsection{Gradient Negation Debiasing Technique}

\begin{wrapfigure}[12]{r}{0.45\linewidth}
\centering
\includegraphics[width=\linewidth]{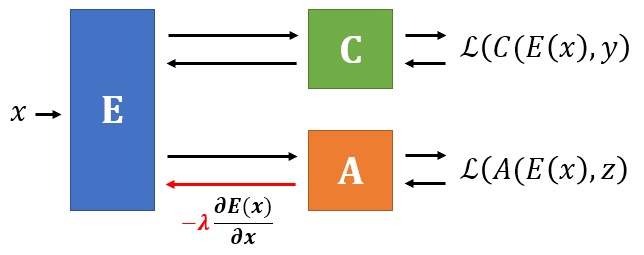}
\caption{High-level training technique for Gradient Negation Debiasing Technique}
\label{fig:fig6}
\end{wrapfigure}

The second debiasing technique we explored was the gradient negation adversarial technique introduced in Beutel 2017 \cite{Beutel}. This technique was originally applied using a very small neural network, and has a similar architecture to that described in Xia 2019 \cite{Xia}, but has a very different training technique. Rather than alternating with different rounds of training for the adversarial technique, both the classifier $C$ and the adversary $A$ take their input from the encoder $E$ and compute the loss with respect to the target attribute $y$ and the sensitive attribute $z$ respectively. However, the difference is in the back-propagation phase (see Figure \ref{fig:fig6}). Namely:

\begin{itemize}
    \item The gradients with respect to the target loss, classifier $C$, and encoder $E$ are applied directly.
    \item The gradients with respect to the sensitive loss and adversary $A$ are applied directly.
    \item 	The gradients with respect to the sensitive loss and encoder $E$ are multiplied by $-\lambda$ before being applied in the back propagation step (thus negating the gradient).
\end{itemize}

This technique allows the target classifier to continue performing well at its task, while the encoder learns to produce output that will “fool” the adversary, much like was proposed in Xia 2019 \cite{Xia}.

\subsection{Additional Training Specifications}

For the purpose of training with the Xia 2019 \cite{Xia} technique, we followed the authors’ suggestion to use a small $\alpha=0.05$ and executed 11 total rounds of training. For the Beutel 2017 \cite{Beutel} technique, we chose small values of $\lambda \in (0, 2]$, experimenting with a few values to see what yields the best performance; using a value of $\lambda=0$ effectively results in a model purely trained for hate speech detection with no debiasing, which we used to train our baseline hate speech model. 

\subsection{Evaluation}
For high level performance evaluations of the hate speech detection task, we concentrated on the traditional performance metrics (such as accuracy, F1). However, given the annotation bias present in the Founta dataset, we also looked to performance metrics such as the per-class false positive rates (FPR) used by both Sap 2019 \cite{Sap} and Xia 2019 \cite{Xia} to evaluate for bias among our two dialects. Additionally, we evaluated overall confusion matricies and confusion matricies per dialect subgroup to compare disparities with and without applying debiasing techniques.

In addition to standard metrics, we examined the fairness metrics described in Beutel 2017 \cite{Beutel}. These metrics depend on the following initial probability definitions (please refer to the original paper for a mathematical definition):

\begin{itemize}
    \item $ProbTrue_{y,z}$ – the probability an example is assigned to the target class $y$, given the example belongs to the subset where the sensitive attribute equals $z$.
    \item $ProbCorrect_{y,z}$ - the probability of correctly identifying an example’s target label as $y$, given the example belongs to the subset where the sensitive attribute equals $z$.
\end{itemize}

With these definitions, we can define the metrics from Beutel 2017 \cite{Beutel} as follows:
$$
ParityGap_y=ProbTrue_{y,z=1}-ProbTrue_{y,z=0}
$$
$$
EqualityGap_y=ProbCorrect_{y,z=1}-ProbCorrect_{y,z=0}
$$

With respect to our hate speech detection task, where our sensitive attribute corresponds to the dialect of the input text, the $ParityGap_y$ effectively measures the disparity in the likelihood for an example to be assigned a particular hate speech label $y$, depending on whether or not the text has an AAE dialect; thus, a parity gap score of zero means equal likelihood among the two dialect subgroups of being assigned to the target class (no bias). By comparison, the $EqualityGap_y$ effectively measures the disparity in the likelihood for an example with a true hate speech label $y$ to be assigned to produce a correct prediction (i.e. recall), depending on whether or not the text has an AAE dialect; thus, an equality gap score of zero means equal likelihood among the two dialect subgroups of doing the correct prediction, given the target class $y$ (no bias).  

\section{Results and Discussion}
We first explored applying both debiasing techniques, using a training set where the hate speech distributions for each dialect subgroup were representative of the original Founta dataset, with equal representation for each dialect subgroup overall, while having representation bias across the two subgroups (refer to Figure \ref{fig:fig2}). 

\begin{figure}[b]
\centering
\includegraphics[width=\linewidth]{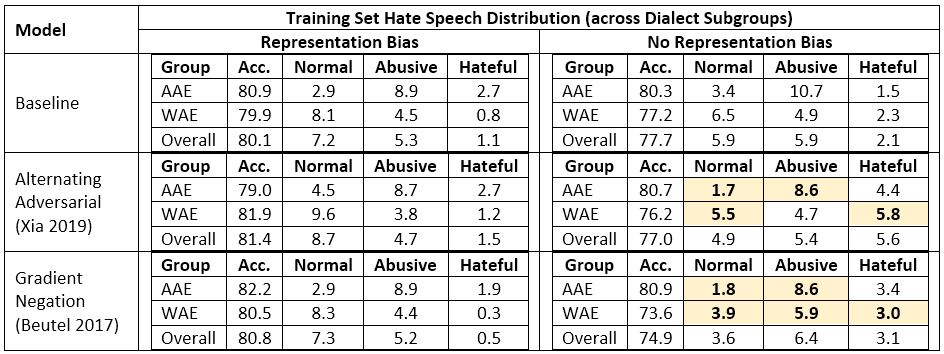}
\caption{Performance results (accuracies and FPRs) for four-class hate speech detection models}
\label{fig:fig7}
\end{figure}

Our baseline model for hate speech detection revealed the same disparity discussed in Sap 2019 \cite{Sap}, as evident in the per-class FPRs shown in Figure \ref{fig:fig7}, where false positives (falsely identifying normal text as abusive/hateful) are more likely for the AAE subgroup and false negatives (falsely identifying abusive/hateful text as normal) more likely for the WAE subgroup. Additionally, the baseline model had an accuracy of 80.1\% on the hate speech detection task, while also having an accuracy of 83\% for the dialect classification task (Appendix Figure \ref{fig:fig14}), again emphasizing the disparity present in the baseline model. 

While the debiasing techniques led to slight improvements in hate speech accuracy overall and did induce poorer performance in the dialect classifier (with most accuracies reducing to less than 35\% on dialect classification (Appendix Figure \ref{fig:fig14}), the techniques seemed to have little-to-no impact on resolving the disparity seen in the per-class FPRs across the two dialect subgroups (Figure \ref{fig:fig7}), as well as the parity and equality metrics (Appendix Figure \ref{fig:fig14}). Even when reducing the four-class problem to a two-class problem (with the positive class as the combination of “abusive” and “hateful”) to force the model to concentrate on the disparities, the debiasing techniques seem to have no effect on resolving the dialect subgroup disparity (Appendix Figures \ref{fig:fig12}, \ref{fig:fig13}, \ref{fig:fig14}).

Considering the discussion in Beutel 2017 \cite{Beutel} around the importance of having equal representation for the target task among our two dialect subgroups, we explored applying the same set of experiments using a training dataset where each dialect subgroup saw the same distribution of hate speech classes (thus eliminating representation bias), while maintaining a balance with respect to dialect. This way, the model was less likely to learn the “default” class distributions (where the original training dataset had more “normal” examples for the WAE subgroup, and more “abusive” examples for the AAE subgroup). These new training set distributions are shown in Figure \ref{fig:fig3}.

After applying the new training set construction strategy, our new baseline model saw a decrease in performance on the hate speech detection task to about 77.7\% accuracy overall on the test set, with approximately the same disparities as in our first set of experiments (high false positives for the AAE subgroup, high false negatives for the WAE subgroup, as well as high performance on the dialect classification task), as shown in the right-hand "no representation bias" side of Figure \ref{fig:fig7}.

\begin{figure}[t]
\centering
\includegraphics[width=\linewidth]{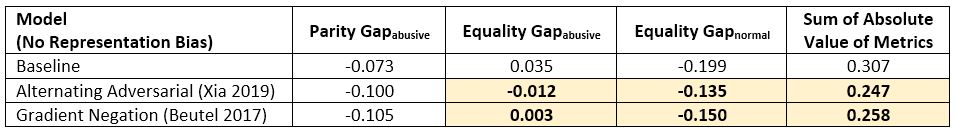}
\caption{Fairness Metrics for Four-Class Hate Speech Models}
\label{fig:fig8}
\end{figure}

However, when we applied the same debiasing techniques as before, the techniques seemed much more effective than in the previous set of experiments. While it’s anticipated that effective debiasing would have resulted in a decrease in false positives for the AAE subgroup (the ideal case), both techniques led to an increase in false positives for the WAE subgroup, alongside a decrease in false negatives for the WAE subgroup (and partially for the AAE subgroup) (Figure \ref{fig:fig7}). Consequently, when comparing fairness metrics (closer to zero being best for all metrics), we see a general improvement in the equality gap metrics as the model begins performing more similarly for both dialect subgroups (Figure \ref{fig:fig8}), thus reducing the disparity. 

The same gains, however, are not observed when reducing the four-class problem to a two-class problem (where the positive class consists of “abusive” and “hateful”) (Appendix Figure \ref{fig:fig13}). This is due in part to the high initial performance on the two-class problem, where the model experiences less confusion on the simplified task and thus, the disparity is not as evident as in the four-class case.

\section{Conclusion}
In this work, we explored applying simple debiasing techniques in the context of transformer-based encoders on the hate speech detection task to evaluate the extent to which these techniques are effective at debiasing these models. We confirmed that models trained on the hate speech task do have a disparity with respect to dialect, propagating the annotation bias present in the original Founta dataset. We conclude that the debiasing techniques are most effective when accounting for both the annotation bias and the representation bias when constructing the model’s training set, thus allowing for an improvement in the disparities over the baseline models.

For future work, we would like to experiment more with increasing the number of rounds of training used on the alternating adversarial technique, as well as experimenting with other data sampling techniques that considers dataset biases, without the need to undersample (to increase the size of the training dataset). We’d also like to explore balancing the training data distributions with respect to hate speech classes, and evaluating the models by applying rigorous probing tasks using advents such as CheckList (Ribeiro 2020 \cite{Ribeiro}) and BertViz (Vig 2019 \cite{Vig}) to better understand the impact of these debiasing techniques on hate speech detection models. 

\section{Acknowledgments}
We would like to thank Su Lin Blodgett, Maarten Sap, and Mengzhou Xia for their help with this work, as well as Joachim Rahmfeld and Mark Butler for advising us throughout this project.

\newpage

\bibliographystyle{unsrt}  
\bibliography{references}

\newpage
\section{Appendix}
\subsection{Detailed Model Architecture Diagrams}

\begin{figure}[H]
\centering
\includegraphics[width=0.95\linewidth]{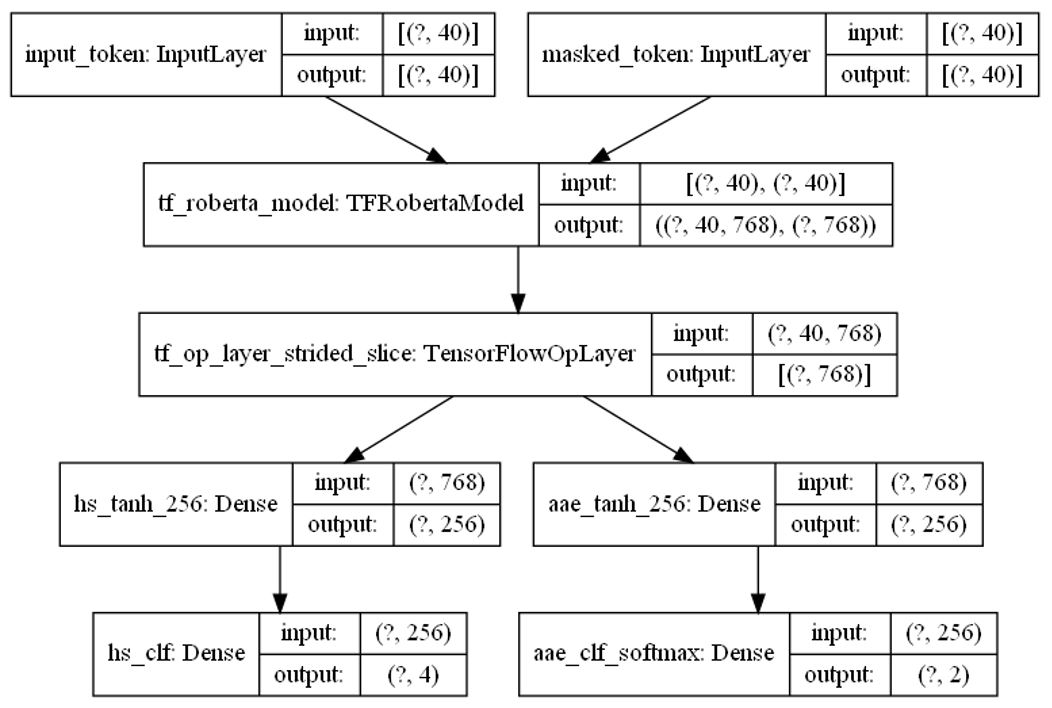}
\caption{Alternating Adversarial Debiasing Model Architecture}
\label{fig:fig9}
\end{figure}

\begin{figure}[H]
\centering
\includegraphics[width=0.95\linewidth]{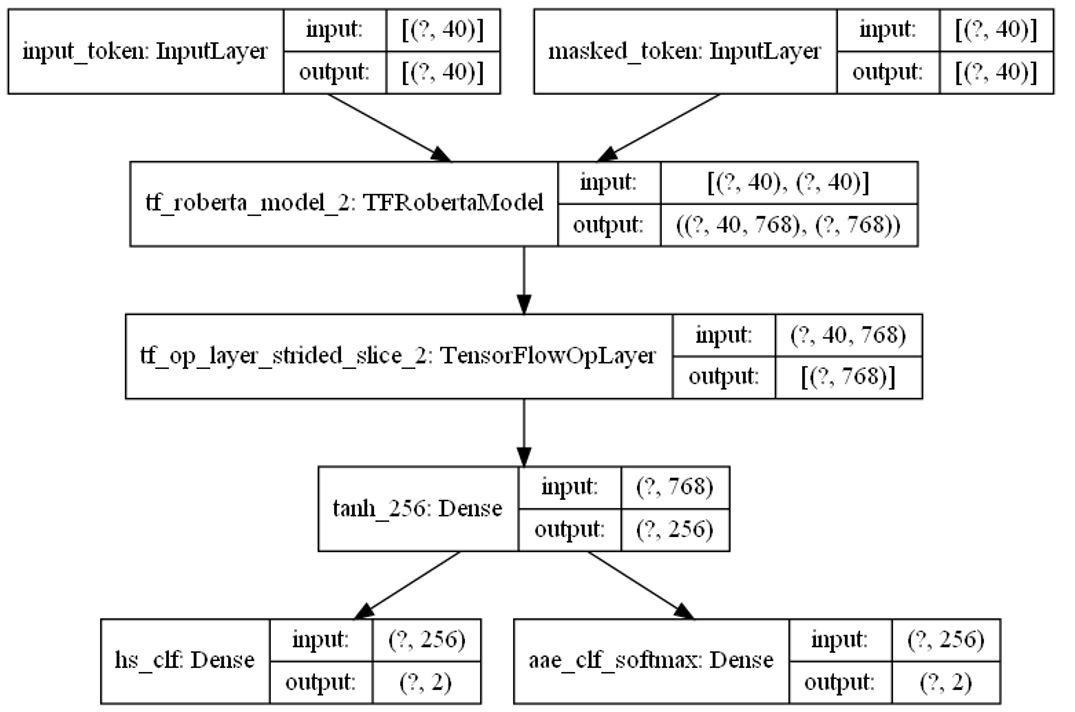}
\caption{Gradient Negation Debiasing Model Architecture}
\label{fig:fig10}
\end{figure}

\subsection{Exploration of Model Prediction Errors}
Given the subset of models where we saw the debiasing techniques have an effect relative to the baseline model, we looked at samples of the subsets of data where the model predictions changed from baseline to debiased model, per dialect subgroup. We list a few observations here about the types of tweets observed that fall into these categories to better understand the nature of errors.

\begin{figure}[ht!]
\centering
\includegraphics[width=\linewidth]{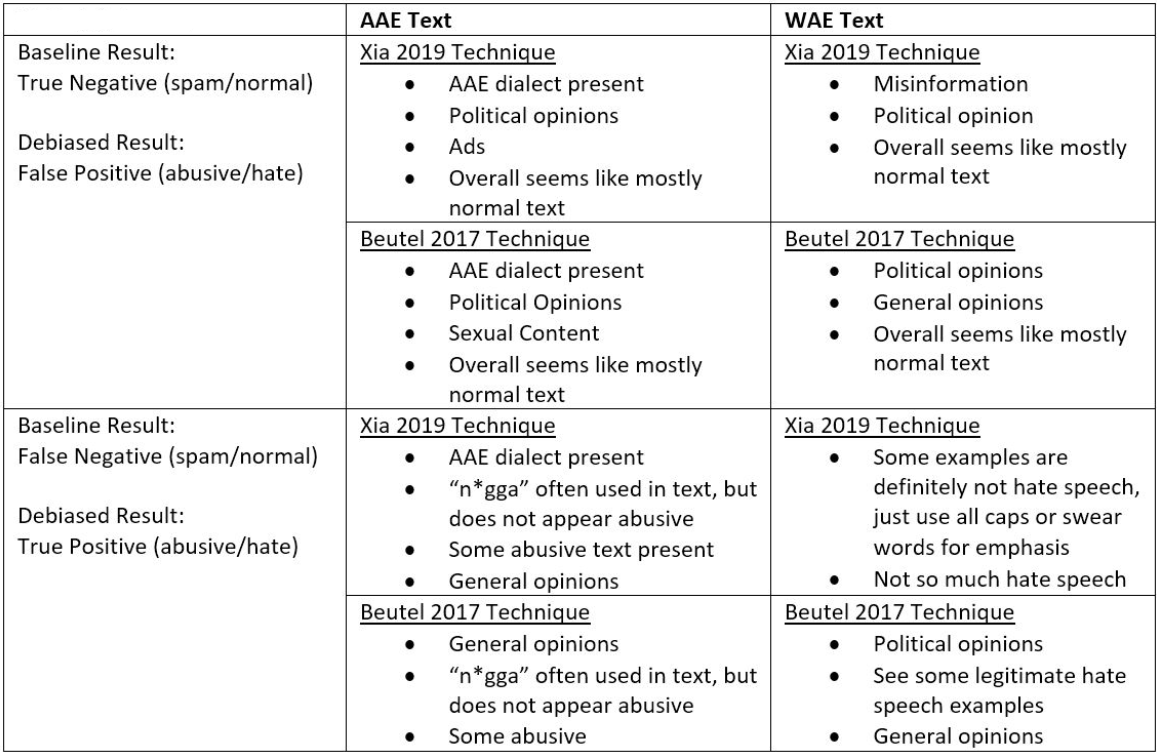}
\caption{General notes on error analysis from hate speech classifier predictions}
\label{fig:fig11}
\end{figure}

\subsection{Per-Class False Positive Rates for Two-Class Hate Speech Detection Models}
Below are the performance results for the two-class based hate speech detection models, which are referenced in section 4. The "Abusive" class below represents the positive class, or the combination of "abusive" and "hate" from the original Founta data. Likewise, the "Normal" class below represents the negative class, or the combination of "normal" and "spam" from the original Founta data.

\begin{figure}[ht!]
\centering
\includegraphics[width=0.93\linewidth]{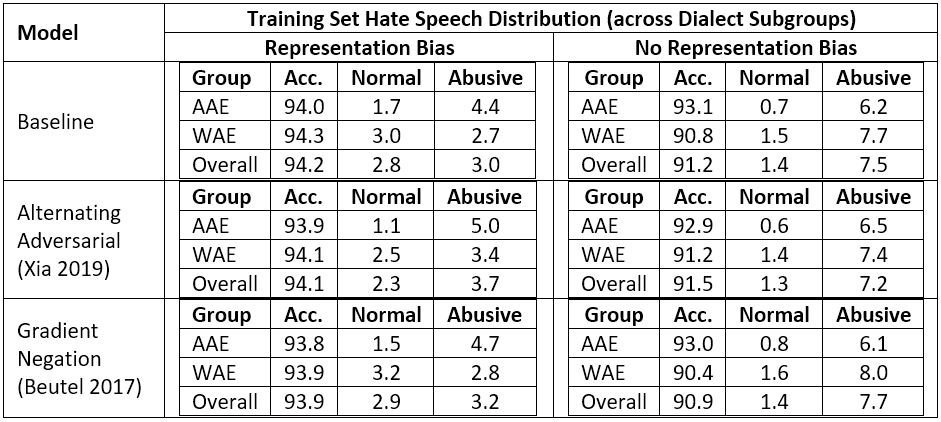}
\caption{Performance results for two-class hate speech detection models}
\label{fig:fig12}
\end{figure}

\newpage

\subsection{Detailed Model Performance Metrics}

Below are detailed model performance metrics for all of the core hate speech classifier models considered in our evaluation. The "\# of Hate Speech Output Classes" corresponds to the situation in which we used all four Founta dataset labels (case = 4) or when we condense them to the two classes of positive ("abusive" and "hateful") and negative ("normal" and "spam") (case = 2).

\begin{figure}[ht!]
\centering
\includegraphics[width=\linewidth]{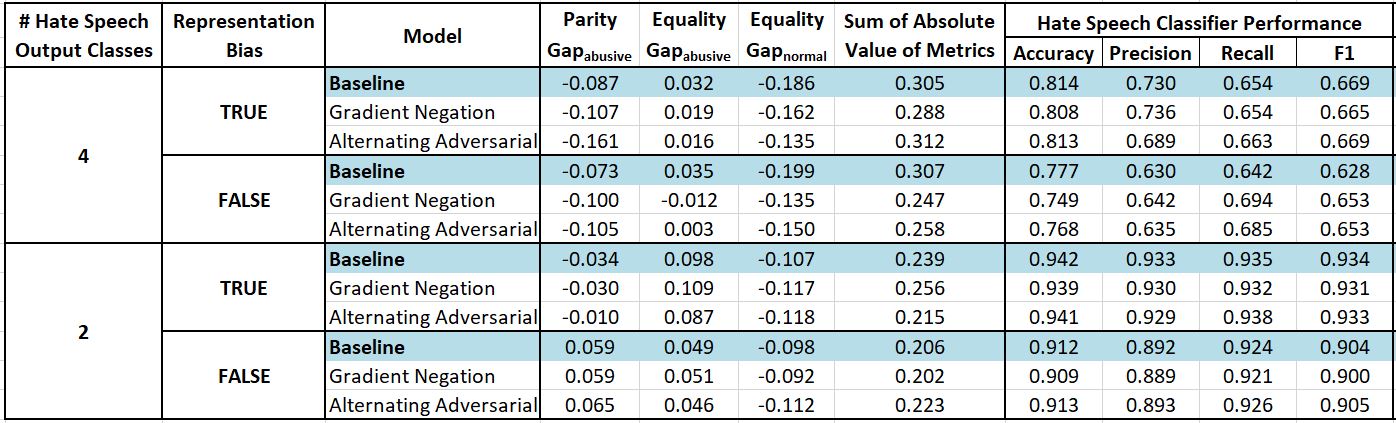}
\caption{Performance metrics for hate speech classifiers}
\label{fig:fig13}
\end{figure}

For each trained encoder used for the hate speech detection task, the same encoder can also be applied as the encoder portion of the dialect classifier. The performance results for these dialect classifiers are shown below. Note that for the second to last model (case where we have two hate speech output classes, no representation bias in the training data, and the debiasing technique is Gradient Negation), we see a very high accuracy of 82.8\%, despite the fact that we are applying a debiasing technique. This dialect classifier, however, always predicts the majority class (WAE) and performs poorly in reality, as seen in the precision, recall, and F1 scores for the same model.

\begin{figure}[ht!]
\centering
\includegraphics[width=\linewidth]{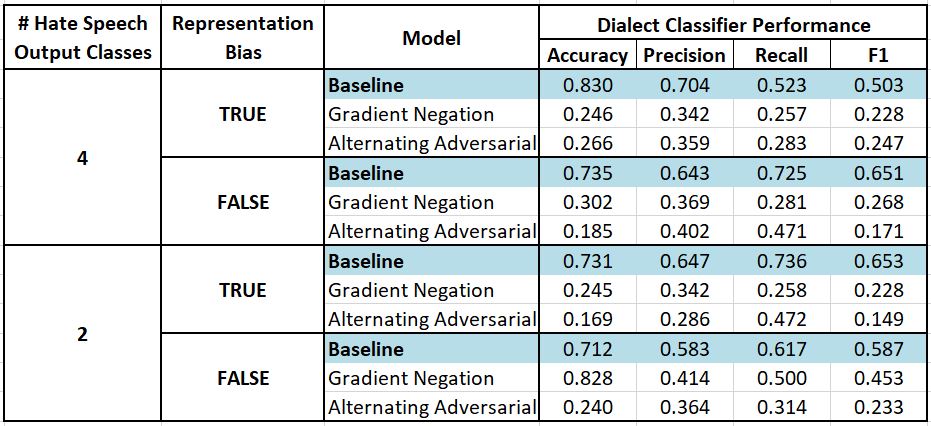}
\caption{Performance metrics for dialect classifiers}
\label{fig:fig14}
\end{figure}

\end{document}